\documentclass[letterpaper]{article}
\usepackage{aaai24}
\usepackage{times}
\usepackage{helvet}
\usepackage{courier}
\usepackage[hyphens]{url}
\usepackage{graphicx}
\usepackage{natbib}
\usepackage{caption}
\usepackage{algorithm}
\usepackage{algorithmic}
\usepackage{multirow}
\usepackage{MnSymbol}
\newcommand\Tstrut{\rule{0pt}{2.3ex}}

\usepackage{newfloat}
\usepackage{listings}
\DeclareCaptionStyle{ruled}{labelfont=normalfont,labelsep=colon,strut=off}
\floatstyle{ruled}
\newfloat{listing}{tb}{lst}{}
\floatname{listing}{Listing}
\title{Augmented Commonsense Knowledge for Remote Object Grounding}

\author{
    Bahram Mohammadi\textsuperscript{\rm 1}, Yicong Hong\textsuperscript{\rm 2}, Yuankai Qi\textsuperscript{\rm 3}, Qi Wu\textsuperscript{\rm 1}, Shirui Pan\textsuperscript{\rm 4}, Javen Qinfeng Shi\textsuperscript{\rm 1} 
}

\affiliations{
    \textsuperscript{\rm 1}Australian Institute for Machine Learning (AIML), University of Adelaide \\
    \textsuperscript{\rm 2}Australian National University \\
    \textsuperscript{\rm 3}Macquarie University \\
    \textsuperscript{\rm 4}Griffith University \\
    
    \{bahram.mohammadi, qi.wu01, javen.shi\}@adelaide.edu.au, \\ yicong.hong@anu.edu.au, yuankai.qi@mq.edu.au, s.pan@griffith.edu.au
}

\begin{document}

\maketitle

\begin{abstract}
The vision-and-language navigation (VLN) task necessitates an agent to perceive the surroundings, follow natural language instructions, and act in photo-realistic unseen environments. Most of the existing methods employ the entire image or object features to represent navigable viewpoints. However, these representations are insufficient for proper action prediction, especially for the REVERIE task, which uses concise high-level instructions, such as ''Bring me the blue cushion in the master bedroom''. To address enhancing representation, we propose an augmented commonsense knowledge model (ACK) to leverage commonsense information as a spatio-temporal knowledge graph for improving agent navigation. Specifically, the proposed approach involves constructing a knowledge base by retrieving commonsense information from ConceptNet, followed by a refinement module to remove noisy and irrelevant knowledge. We further present ACK which consists of knowledge graph-aware cross-modal and concept aggregation modules to enhance visual representation and visual-textual data alignment by integrating visible objects, commonsense knowledge, and concept history, which includes object and knowledge temporal information. Moreover, we add a new pipeline for the commonsense-based decision-making process which leads to more accurate local action prediction. Experimental results demonstrate our proposed model noticeably outperforms the baseline and archives the state-of-the-art on the REVERIE benchmark. The source code is available at https://github.com/Bahram-Mohammadi/ACK.
\end{abstract}

\begin{figure}[t]
\begin{center}
\includegraphics[width=0.99\linewidth]{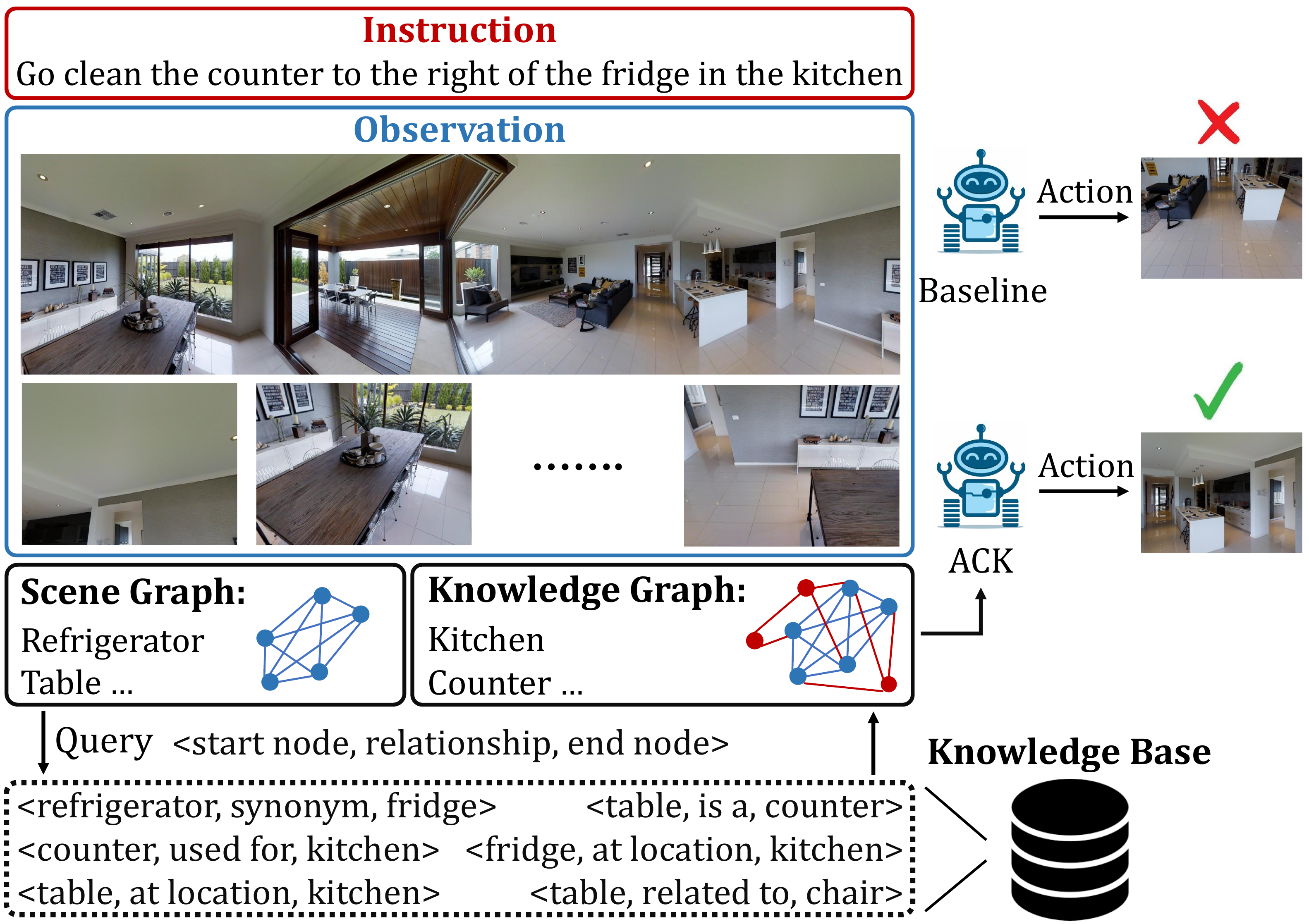}
\end{center}
   \caption{Action prediction by the baseline and our method. Utilizing visible objects alongside commonsense knowledge as a spatio-temporal knowledge graph improves visual representation and action prediction. Best viewed in color.}
\label{fig:teaser}
\end{figure}

\section{Introduction}
Navigating an embodied agent through complex and unseen environments by following natural language instructions is a challenging problem in artificial intelligence research. In this regard, vision-and-language navigation (VLN)~\cite{Anderson_2018_CVPR} has drawn the attention of many researchers in recent years~\cite{Fried_2018_speaker, tan_2019_envdrop, Hao_2020_prevalent, Zhu_2021_gbe, Chen_2022_duet, li_2023_kerm} and a variety of VLN tasks have been introduced in different levels, such as room-to-room (R2R)~\cite{Anderson_2018_CVPR} and remote embodied visual referring expression in real indoor environments (REVERIE)~\cite{Qi_2020_reverie}. In R2R, the agent aims to reach a pre-specified location from a starting point by following fine-grained instructions. However, To be more practical, REVERIE introduces a goal-oriented task in which the agent needs to explore the environment and localize the target object according to concise instructions, e.g., "\textit{Go to the living room and clean the table next to the couch}". Therefore, the agent cannot complete the task successfully just by strictly following instructions and requires more information about the environment to predict the correct action at each step.

Many of the proposed methods exploit scene-level features to represent visual perception \cite{qiao_2022_hop, Guhur_2021_airbert, Shizhe_2021_hamt}. To provide the agent with richer visual clues, a wide range of previous works \cite{Hong_2020_Nurips, Qi_2020_Object, Qi_2021_ICCV} use object-level features. It is very common to use an object as a landmark in the instructions which means they can also be utilized as visual landmarks. Hence, object localization along with their orientation encoding regarding the heading and elevation angles of the agent is very beneficial to align the detected objects in the scene with the object labels in the instruction. For example, Given the instruction "\textit{With the fireplace on your right walk down the walkway and stop at the end before you enter the next room}", the agent first needs to detect the fireplace and then correctly identify it as the visual landmark on the right to predict the proper action. High-level instructions and the lack of sufficient data in REVERIE disturb the navigation performance. This problem motivated us to provide the agent with additional information that is not present in the scene but can be inferred from the visible content of navigable directions. As Humans need to reason over complementary information to make a more precise decision, we aim to introduce commonsense knowledge into the REVERIE task to improve navigation. Employing external knowledge has shown significant performance in other vision-and-language problems such as image captioning and visual question answering (VQA) \cite{Marino_2019_okvqa, Marino_2021_krisp, Ding_2022_mukea}. Not only does such information enhance the vision and text alignment, but it also generalizes the action reasoning of the agent. Furthermore, the spatial and temporal connection between objects and knowledge during navigation boosts the exploration and generalization ability of the agent in unseen environments. For instance, \textit{hallway} is followed by \textit{living room} in the training environment, thus learning this sequence helps the agent to take the right action by observing the same concept, \textit{hallway}, in an unseen environment. 

In this work, to fulfill the above-mentioned requirements, we incorporate commonsense knowledge into the REVERIE task as a spatio-temporal knowledge graph by constructing a knowledge base followed by our proposed method ACK. We build an internal knowledge base according to the visible contents of the image and ConceptNet~\cite{liu_2004_conceptnet} as the external knowledge base. Then, the pre-trained contrastive language-image pre-training (CLIP) model~\cite{radford_2021_Learning} is utilized to collect and rank the most pertinent knowledge to the scene and detected objects. These object and knowledge features are complementary to the existing visual representation and align the viewpoint images with instructions. Afterward, we present ACK which takes advantage of commonsense knowledge to enhance visual representation and action reasoning. ACK consists of two modules to integrate object and knowledge features along with their historical information. The knowledge graph-aware cross-modal encoder models the relationship between concepts and instructions while the concept history of the previous step is taken into account. Subsequently, the concept aggregation module outputs a single concept feature per each navigable direction which is utilized for visual representation enhancement and local action prediction in the commonsense-based decision-making pipeline. Figure~\ref{fig:teaser} demonstrates the action prediction comparison between DUET \cite{Chen_2022_duet} as the baseline and ACK. As shown in this figure, our model can make the right decision by exploiting object-level and knowledge-level features.

The experiments are conducted on the REVERIE dataset and results show that our proposed approach, ACK, outperforms the state-of-the-art methods. In summary, the main contributions of this work are as follows:

\begin{itemize}
    \item 
    We integrate object-level features with object-related commonsense knowledge to complement the existing visual representation and enhance vision-text alignment. 
        
    \item
    We propose ACK followed by the commonsense-based decision-making pipeline for leveraging object and knowledge features to provide a more informative representation and hold their historical information for making a sequence of correct decisions.
 
    \item
    We conduct experiments on REVERIE to validate the effectiveness and generalization ability of ACK and the results show the superiority of our approach over the state-of-the-art methods and the baseline model.
\end{itemize}

\begin{figure*}[!t]
\begin{center}
\includegraphics[width=0.99\linewidth]{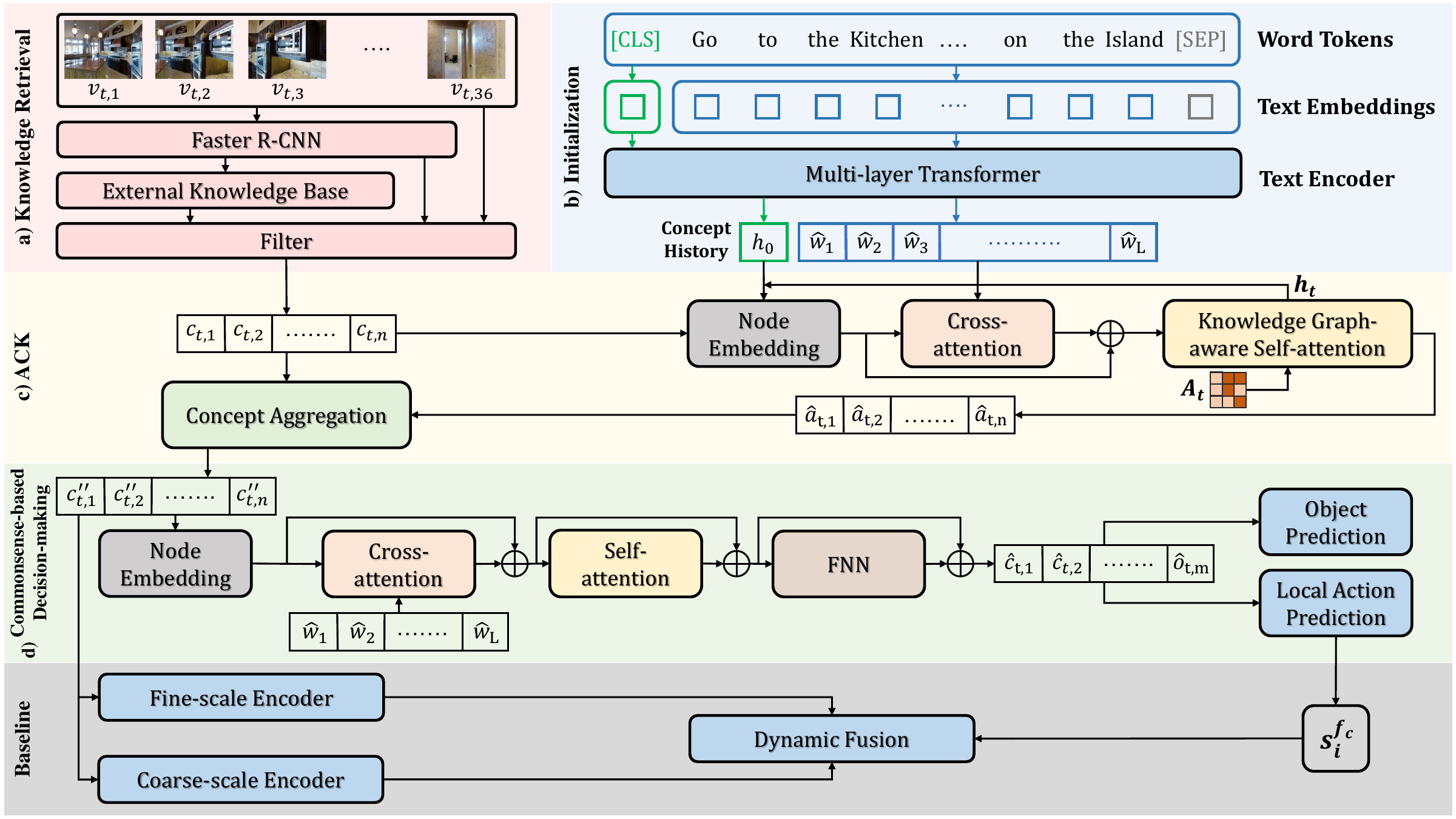}
\end{center}
   \caption{Main architecture of our proposed method. (a) Retrieving and refining the commonsense knowledge. (b) Initializing the concept history which represents the entire instruction and obtaining the text embedding. (c) ACK receives the detected objects, commonsense knowledge, and their temporal information to output weighted raw concept features which are utilized in the commonsense-based decision-making pipeline and the baseline model. (d) Inspired by the baseline agent, we add a new pipeline to produce the local action score and predict the object. Best viewed in color.}
\label{fig:overview}
\end{figure*}

\section{Related Work}
\noindent\textbf{Vision-and-Language Navigation.}
In the VLN task, an agent is required to find the optimal path toward the target location given the visual and textual input data. In recent years, numerous approaches ~\cite{Wang_2021_ssm, Deng_2020_Evolving, Lin_2022_adapt, Liu_2021_mixup} have been proposed to improve the performance of the agent. The first baseline for the VLN is introduced by~\cite{Anderson_2018_CVPR} which designs a multi-modal Seq2Seq model. Then~\cite{Fried_2018_speaker} proposes the speaker-follower method to augment the data and improve the generalization. EnvDrop~\cite{tan_2019_envdrop} extends this work by presenting environmental dropout. \cite{Wang_2019_reinforced} presents a module for cross-modal grounding which enables the agent to infer the essential parts of the scenes and sub-instructions. \cite{ma_2019_smna} focuses on progress monitoring regarding the instruction. In light of the successes of vision-and-language pre-training PREVALENT~\cite{Hao_2020_prevalent} pre-trains the navigation model using the self-learning approach and AirBERT~\cite{Guhur_2021_airbert} improves cross-modality interaction. Many of the recent works in VLN rely on the navigation history to achieve superior performance. Some approaches condense the history into a single vector~\cite{Wang_2019_reinforced, Hong_2021_recbert} while others attempt to explicitly store the previous states \cite{Pashevich_2021_episodic, Chen_2022_duet}. Most recently, \cite{gap} propose to especially mind the passed target location in trajectory histories.

\noindent\textbf{VLN with Commonsense Knowledge.}
In the context of VLN, incorporating commonsense knowledge is rarely considered in previous works, however, it has recently drawn the attention of researchers in this topic~\cite{Gao_2021_CVPR, li_2022_incorporating, li_2023_kerm}. The most widely-used large-scale structured knowledge bases are ConceptNet~\cite{liu_2004_conceptnet} and DBpedia~\cite{auer_2007_dbpedia} which are created by automatic data extraction and manual annotation, respectively. CKR~\cite{Gao_2021_CVPR} exploits ConceptNet to iteratively perform object- and room-entity reasoning through internal and external graph reasoning during the training. In another line of work, KERM~\cite{li_2023_kerm} proposes a knowledge-enhanced reasoning model to take advantage of external knowledge which is retrieved from the visual genome~\cite{krishna_2017_visual} by parsing the region descriptions. In this work, we leverage commonsense knowledge to properly align viewpoints with instructions and improve visual representation as well as local action prediction. The architecture of DUET~\cite{Chen_2022_duet} is followed as the main baseline of our proposed method.

\section{Method}
In this section, we first explain the problem formulation and the overview of our proposed approach. Then, we present commonsense knowledge retrieval and ACK in detail and describe the commonsense-based decision-making pipeline for local action prediction.

\noindent\textbf{Problem Formulation.} 
In the context of REVERIE~\cite{Qi_2020_reverie}, given a language instruction as a sequence of words denoted as $I = \{w_i\}^{L}_{i=1}$, where $w_{i}$ represents the $i^{th}$ word and $L$ is the length of the sequence, the agent navigates through an undirected connectivity graph $\mathcal{G} = \{\mathcal{V}, \mathcal{E}\}$, where $\mathcal{V}$ corresponds to viewpoints and $\mathcal{E}$ shows the connection between nodes. At time step $t$, the agent is located at node $\mathcal{V}_{t}$ and perceives a panoramic view $\mathcal{V}_{t} = \{v_{t, i}\}^{36}_{i=1}$. The agent infers an action $a_{t}$ to transfer from state $s_{t}$ to state $s_{t+1}$ only based on the navigable directions $\mathcal{N}(\mathcal{V}_{t}) = \{v_{t, i}\}^{K}_{i=1}$, where $\mathcal{N}(\mathcal{V}_{t}) \subseteq \mathcal{V}_{t}$. Each state includes a triplet $\{v_{t, i}, \theta_{t, i}, \psi_{t, i}\}$, where $v_{t, i}$ is the viewpoint image, and $\{\theta_{t, i}, \psi_{t, i}\}$ are angels of heading and elevation, respectively, to determine the orientation of the image with respect to the agent. The agent is required to walk on the connectivity graph by selecting the next location at each node until it decides to stop or the number of action steps exceeds the threshold. The episode ends when the agent identifies the position of the target object within the panoramic view.

\noindent\textbf{Method Overview.}
We follow DUET\cite{Chen_2022_duet} architecture as the main baseline. This method includes two modules, topological mapping and global action planning. The former module is responsible for the gradual construction of a map over time by adding new observed locations during the path and updating the representation of each node. Afterward, the action, including the next location or stop action, is predicted by the latter module. DUET dynamically fuses action prediction of two scales: a fine-scale representation of the current location and a coarse-scale representation of the topological map to balance fine-grained language grounding against reasoning over the graphs.

In this work, we aim to improve the visual representation, which affects both fine- and coarse-scale encoders, and local action prediction. To do so, we incorporate commonsense knowledge into the baseline agent by leveraging the visible entities in viewpoint images and adding a new pipeline for local action reasoning. Through this paper, we use concepts and objects/knowledge interchangeably. To achieve more accurate alignment between instructions and candidate directions, we utilize object labels rather than their bounding boxes. As shown in Figure~\ref{fig:overview}(a) we first build a knowledge base using detected objects and an external knowledge base followed by a filtering module to refine it by removing noisy and irrelevant data according to the entire image and objects. In the initialization phase Figure~\ref{fig:overview}(b), we obtain the text embedding and initialize concept history as a representation of the entire instruction. Then, we exploit ACK, Figure~\ref{fig:overview}(c), to generate weighted raw concept features. Finally, we add a new pipeline to reason over the concept features and output the scores for object and local action prediction as illustrated in Figure~\ref{fig:overview}(d). These scores are used as inputs for dynamic fusion to help the agent predict the correct local action.

\subsection{Commonsense Knowledge Retrieval}
Inspired by the human decision-making process which is based on background knowledge, we aim to provide the agent with complementary information to scene-level and object-level data. Commonsense knowledge not only helps the agent to understand the surroundings comprehensively, but it also facilitates data matching between the images and instructions. To obtain appropriate external knowledge, we first construct a knowledge base and then filter out the irrelevant information regarding visible content in each direction.

\noindent\textbf{Object Detection.} For $v_{t, i}$ which is the $i^{th}$ view at time step $t$, the Faster R-CNN model \cite{shaoqing_2015_faster} pre-trained on visual genome (VG) \cite{krishna_2017_visual} is utilized to obtain the object-set $\mathcal{O}_{v_{t, i}}$. To avoid overlooking helpful information we use all of the objects not just the most salient ones. During navigation, the adopted object detector is capable of differentiating 1600 categories $\{o_{i}\}^{1600}_{i=1}$, including those that have been annotated in the REVERIE dataset.

\noindent\textbf{Knowledge Base Construction.}
To build the internal knowledge base, we employ ConceptNet~\cite{liu_2004_conceptnet} as the external source of information. Each query we send to ConceptNet contains three parameters, start node, end node, and relationship type. The response from the ConceptNet is represented by a tuple $f_{i, j} = (s_{i}, r_{i, j}, e_{j})$, which indicates that the start node $s_{i}$ is connected to the end node $e_{j}$ through the relationship $r_{i, j}$. More than 30 different relationships are available in ConceptNet, but we only use 8 most relevant ones based on their descriptions\footnote{https://github.com/commonsense/conceptnet5/wiki/Relations}. For each object set $\mathcal{O}_{v_{t, i}}$, all related data is extracted from the knowledge base. For example, if $\mathcal{O}_{v_{t, i}}$ contains \textit{bed}, then $\mathcal{K}_{v_{t, i}}$ includes \textit{bedroom} according to the triplet (\textit{bed}, \textit{AtLocation}, \textit{bedroom}) and the agent can infer it faces the \textit{bedroom}.

\noindent\textbf{Knowledge Selection.}
The extracted knowledge may be irrelevant or noisy, which could potentially impact the accuracy of the downstream task. To address this issue, we consider a refinement module to select the top-$k$ pertinent supporting facts for viewpoints $\{k_{t, i}\}^{K}_{i=1}$. To do so, the pre-trained CLIP model~\cite{radford_2021_Learning} is used which consists of CLIP-I and CLIP-T encoders for encoding image and text, respectively, into a joint embedding space. We employ CLIP-I to encode entire images while object and knowledge labels are encoded by CLIP-T. Afterward, we calculate the similarity score for each fact according to its average cosine similarity with the whole image and its objects. The higher score means the corresponding knowledge is more suitable to be utilized. Finally, we select top-$k$ commonsense knowledge. In this case, even for the same object sets we may retrieve different knowledge sets.

\subsection{Augmented Commonsense Knowledge Model}
At time step $t$, the object features $\mathcal{O}_t$, knowledge features $\mathcal{K}_t$, concept history features $h_{t-1}$, and instruction features $\mathcal{I}_t$ are fed into our proposed model to either improve visual representation or local action prediction. 

\subsubsection{Knowledge Graph}
This study intends to treat objects, knowledge, and their history as a spatio-temporal knowledge graph. We first generate a fully-connected scene graph using detected objects and then expand it to construct the knowledge graph by adding extracted knowledge. To provide the agent with temporal insights about concepts during navigation, we consider an extra node that is connected to all the others. In the lines below, we elucidate the node representation and adjacency matrix generation.

\begin{figure}[!t]
\begin{center}
\includegraphics[width=0.90\linewidth]{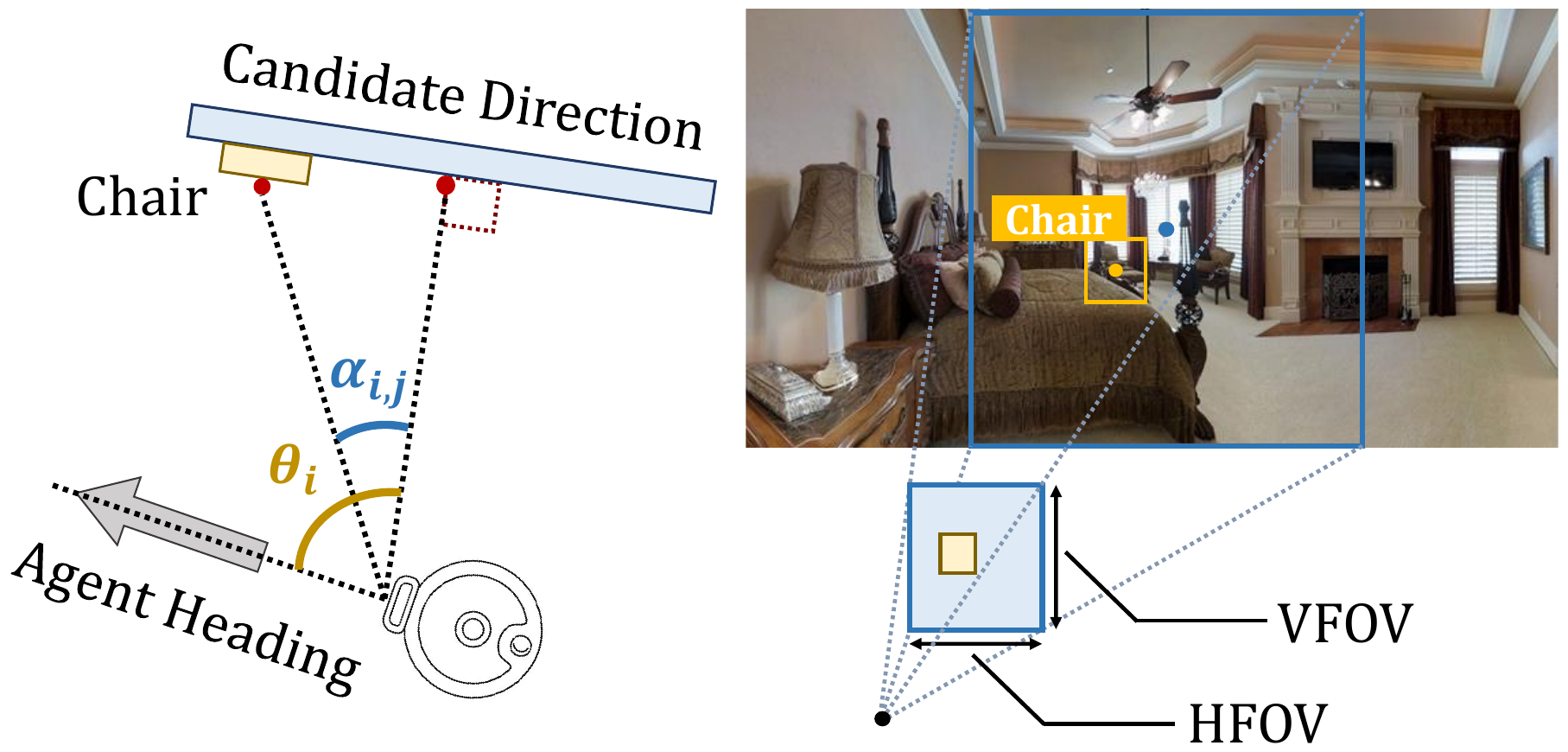}
\end{center}
\caption{Encoding the relative position of objects with respect to the heading and elevation angles of the agent.}
\label{fig:object_position}
\end{figure}

\noindent\textbf{Node Embedding.}
The acquired knowledge exists in textual format, thus, in order to maintain consistency between objects and knowledge and also to provide a better vision-text alignment, object labels are used instead of their visual features. To encode the object and knowledge labels CLIP-T is used. Node type encoding and directional encoding are added to each node feature. The node type encoding embeds 0 for history, 1 for objects, and 2 for knowledge to distinguish between different nodes. The directional encoding embeds the relative position of objects with respect to the heading and elevation angles of the agent. We aim to use objects as visual landmarks to improve alignment between textual and visual data. As shown in Figure~\ref{fig:object_position}, the relative orientation of objects is encoded as follows:

\begin{equation}
    \begin{split}
        \{\mathcal{D}_{t, \mathcal{O}_{i, j}}\}^{N}_{j=1} = & (\sin{(\theta_{i} \pm \alpha_{i, j})}, \cos{(\theta_{i} \pm \beta_{i, j})}, \\
        & \sin{(\psi_{i} \pm \alpha^{\prime}_{i, j})}, \cos{(\psi_{i} \pm \beta^{\prime}_{i, j})})
    \end{split}
\end{equation}

\noindent where $\mathcal{D}_{t, \mathcal{O}_{i, j}}$ is the directional encoding of $j^{th}$ object in the viewpoint $v_{i}$ at time step $t$ and $N$ is the number of objects. The directional encoding for knowledge is set to zero $\{\mathcal{D}_{t, \mathcal{K}_{i, j}}\}^{K}_{j=1} = \vec{0}$, where $K$ is the number of knowledge.

\noindent\textbf{Adjacency Matrix.} At time step $t$, we generate a fully-connected scene graph for each navigable direction $v_{t, i}$ using the detected object set $\mathcal{O}_{v_{t, i}}$. We then add knowledge nodes and establish the connections between concepts if they are correlated in the knowledge base. Specifically, we connect two nodes that represent either objects or knowledge if their relationship is defined in the knowledge base as the triplet $(c_{i}, r_{i,j}, c_{j})$, where $c_{i}$ and $c_{j}$ denote concept nodes and $r_{i,j}$ denotes the relationship between them. The connections between nodes are considered undirected.

\subsubsection{History Initialization}
It is not necessary to encode instructions during the navigation $t>0$. Therefore, to encode the entire instructions at the initialization step $t=0$, they are fed to a multi-layer transformer as the text encoder. The embedded \texttt{[CLS]} token is defined as the initial representation of concept history $h_0$ that represents the entire instruction.

\begin{equation}
    h_0,~\hat{\mathcal{I}} = \text{TextEncoder}(\texttt{[CLS]}, I, \texttt{[SEP]})
\end{equation}

\noindent where \texttt{[CLS]} and \texttt{[SEP]} are pre-defined tokens in BERT model~\cite{devlin_2019_bert} and $I$ is the language instruction.

\subsubsection{Knowledge Graph-aware Cross-modal Encoder}
Having obtained concept features $[\mathcal{C}_t]$, including object and knowledge features $[\mathcal{O}_t;\mathcal{K}_t]$, and the history feature of previous time step $h_{t-1}$, we use a multi-layer cross-modal transformer to model the interaction between concepts, history, and instruction. Each layer of this transformer consists of a cross-attention layer and a self-attention layer. In the cross-attention layer, the concatenation of concept and history features $[h_{t-1};\mathcal{C}_t]$ serves as queries while instruction embedding $\hat{\mathcal{I}}$ is fed to the transformer as keys and values. The cross-attention between $[h_{t-1};\mathcal{C}_t]$ and $\hat{\mathcal{I}}$ is calculated as:

\begin{equation}
    [h^{'}_{t-1};\mathcal{C}^{'}_{t}] = \text{CrossAttention}([h_{t-1};\mathcal{C}_t], \hat{\mathcal{I}}_{t})
\end{equation}

In general, transformers do not consider the structure of the input features. To address this problem, we introduce a knowledge graph-aware self-attention that slightly differs from the standard attention mechanism in standard transformers. In this case, the constructed knowledge graph structure is exploited to compute the attention as follows:

\begin{equation}
    \begin{split}
        [h_{t};\mathcal{A}_{t,\mathcal{C}} & ] = \text{KGS}([h^{'}_{t-1};\mathcal{C}^{'}_t]) \\
        \text{KGS}(X) = \text{Softmax} & \left( \frac{XW_{q}(XW_{k})^{T}}{\sqrt{d}} + M \right) XW_{v} \\
        & M = A_{t}W_{a} + b_{a} \\
    \end{split}
    \label{eq:kgs}
\end{equation}

\noindent where $h_{t}$ is the concept history of the current step, $\mathcal{A}_{t,\mathcal{C}}$ is concept self-attention scores, $X$ is the node representation $[h_{t-1};\mathcal{C}_t]$, $A$ is the adjacency matrix, and $W_a$, $b_a$ are two learnable parameters.

\begin{table*}[t]
  \Large
  \begin{center}
  \resizebox{\textwidth}{!}{
  \begin{tabular}{l|cccc|cc|cccc|cc}
    \hline \hline
    \multicolumn{1}{c|}{\multirow{3}{*}{Methods}} & \multicolumn{6}{c|}{Validation Unseen} & \multicolumn{6}{c}{Test Unseen} \Tstrut \\
    \cline{2-13} & \multicolumn{4}{c|}{Navigation} & \multicolumn{2}{c|}{Grounding} & \multicolumn{4}{c|}{Navigation} & \multicolumn{2}{c}{Grounding}  \\
    \cline{2-13} & 
    \multicolumn{1}{c}{TL} & \multicolumn{1}{c}{OSR$\uparrow$} &
    \multicolumn{1}{c}{SR$\uparrow$} & \multicolumn{1}{c|}{SPL$\uparrow$} & \multicolumn{1}{c}{RGS$\uparrow$} & \multicolumn{1}{c|}{RGSPL$\uparrow$} & \multicolumn{1}{c}{TL} & \multicolumn{1}{c}{OSR$\uparrow$} &
    \multicolumn{1}{c}{SR$\uparrow$} & \multicolumn{1}{c|}{SPL$\uparrow$} & \multicolumn{1}{c}{RGS$\uparrow$} & \multicolumn{1}{c}{RGSPL$\uparrow$} \Tstrut\\
    \hline \hline

    Human & - & - & - & - & - & - & 21.18 & 86.83 & 81.51 & 53.66 & 77.84 & 51.44  \\

    \hline

    Seq2Seq~\cite{Anderson_2018_CVPR} & 11.07 & 8.07 & 4.20 & 2.84 & 2.16 & 1.63 & 10.89 & 6.88 & 3.99 & 3.09 & 2.00 & 1.58  \\

    SMNA~\cite{ma_2019_smna} & 9.07 & 11.28 & 8.15 & 6.44 & 4.54 & 3.61 & 9.23 & 8.39 & 5.80 & 4.53 & 3.10 & 2.39  \\

    RCM~\cite{Wang_2019_reinforced} & 11.98 & 14.23 & 9.29 & 6.97 & 4.89 & 3.89 & 10.60 & 11.68 & 7.84 & 6.67 & 3.67 & 3.14  \\

    FAST-MATTN~\cite{Qi_2020_reverie} & 45.28 & 28.20 & 14.40 & 7.19 & 7.84 & 4.67 & 39.05 & 30.63 & 19.88 & 11.61 & 11.28 & 6.08  \\

    CKR~\cite{Gao_2021_CVPR} & 26.26 & 31.44 & 19.14 & 11.84 & 11.45 & - & 22.46 & 30.40 & 22.00 & 14.25 & 11.60 & - \\

    SIA~\cite{Lin_2021_scene} & 41.53 & 44.67 & 31.53 & 16.28 & 22.41 & 11.56 & 48.61 & 44.56 & 30.80 & 14.85 & 19.02 & 9.20  \\

    ORIST~\cite{Qi_2021_ICCV} & 10.90 & 25.02 & 16.84 & 15.14 & 8.52 & 7.58 & 11.38 & 29.20 & 22.19 & 18.97 & 10.68 & 9.28 \\

    Airbert~\cite{Guhur_2021_airbert} & 18.71 & 34.51 & 27.89 & 21.88 & 18.23 & 14.18 & 17.91 & 34.20 & 30.28 & 23.61 & 16.83 & 13.28  \\

    RecBERT~\cite{Hong_2021_recbert} & 16.78 & 35.02 & 30.67 & 24.90 & 18.77 & 15.27 & 15.86 & 32.91 & 29.61 & 23.99 & 16.50 & 13.51  \\

    HOP~\cite{qiao_2022_hop} & 16.46 & 36.24 & 31.78 & 26.11 & 18.85 & 15.73 & 16.38 & 33.06 & 30.17 & 24.34 & 17.69 & 14.34  \\

    HAMT~\cite{Shizhe_2021_hamt} & 14.08 & 36.84 & 32.95 & 30.20 & 18.92 & 17.28 & 13.62 & 33.41 & 30.40 & 26.67 & 14.88 & 13.08  \\

    TD-STP~\cite{zhao_2022_target} & - & 39.48 & 34.88 & 27.32 & 21.16 & 16.56 & - & 40.26 & 35.89 & 27.51 & 19.88 & 15.40 \\

    AZHP~\cite{gao_2023_azhp} & 22.32 & \textbf{53.65} & 48.31 & \textbf{36.63} & \textbf{34.00} & \textbf{25.79} & 21.84 & 55.31 & 51.57 & 35.85 & 32.25 & 22.44  \\

    KERM~\cite{li_2023_kerm} & 22.47 & \textbf{53.65} & \textbf{49.02} & 34.83 & 33.97 & 24.14 & 18.38 & 57.44 & 52.26 & 37.46 & 32.69 & \textbf{23.15}  \\

    \hline

    DUET~\cite{Chen_2022_duet} & 22.11 & 51.07 & 46.98 & 33.73 & 32.15 & 23.03 & 21.30 & 56.91 & 52.51 & 36.06 & 31.88 & 22.06  \\

    ACK (Ours) & 22.86 & 52.77 &47.49 & 34.44 & 32.66 & 23.92 & 20.65 & \textbf{59.01} & \textbf{53.97} & \textbf{37.89} & \textbf{32.77} & \textbf{23.15}  \\
    
    \hline \hline
  \end{tabular}}
\end{center}
\caption{Comparison of the agent performance with state-of-the-art methods on REVERIE dataset in the single-run setting.}
\label{tab:reverie_results}
\end{table*}

\subsubsection{Concept Aggregation}
At time step $t$, $\mathcal{C}^{k}_{t}$ is the concept tokens at head $k$ and $\mathcal{A}^{k}_{t,\mathcal{C}}$ is the attention scores over the concept tokens. Then, we average the score over all the attention heads ($K=12$) and apply a Softmax function to get the overall concept attention weights as:

\begin{equation}
    \hat{\mathcal{A}}_{t,\mathcal{C}} = \text{Softmax}(\Bar{\mathcal{A}}_{t,\mathcal{C}}) = \text{Softmax} \left( \frac{1}{K} \sum\limits^{K}_{k=1} \mathcal{A}^{k}_{t,\mathcal{C}} \right)
\end{equation}

Now, we perform a weighted sum over the input concept tokens to retrieve the weighted raw concept features as:

\begin{equation}
    \mathcal{C}^{''}_t = \hat{\mathcal{A}}_{t,\mathcal{C}} ~ \mathcal{C}_t
\end{equation}

\noindent where $\mathcal{C}^{''}_t$ is the aggregated concept features since we need one concept feature per each viewpoint image.

\subsubsection{Commonsense-based Decision-making Pipeline}
Inspired by the baseline agent, we add a new commonsense-based decision-making pipeline to predict a navigation score $s^{f_{\mathcal{C}}}$ and an object score $s^{\mathcal{O}}$ to enhance local action reasoning. Afterward, these scores are used in the dynamic fusion module of the baseline model to calculate the final score and predict the next action.

\subsubsection{Training and Inference}
We pre-train the baseline model on single-step action prediction (SAP)~\cite{krantz_2020_beyond}, masked language modeling (MLM)~\cite{devlin_2019_bert}, masked region classification (MRC)~\cite{lu_2019_vilbert}, and object grounding (OG)~\cite{Lin_2021_scene} tasks. However, our proposed model is only incorporated into policy learning. Analogous to the baseline, in addition to SAP loss $L_{SAP}$ and OG loss $L_{OG}$, fine-tuning is guided by supervision provided by a pseudo-interactive demonstrator instead of behavioral cloning. In this case, the agent selects the next location with the overall shortest distance to the destination. We also use the loss calculated through the commonsense-based decision-making pipeline $L_{CD}$. 

For inference, an action is predicted by the agent in each time step. If the agent exceeds the maximum number of action steps or the predicted action is the stop action, it stops at the current node. Otherwise, the agent moves to the predicted state. Eventually, when the agent stops at the final location, the object with the highest score is selected as the designated target object.

\section{Experiments}
\subsection{Implementation Details}
The ACK is not incorporated into pre-training tasks of DUET~\cite{Chen_2022_duet} and we only fine-tune the proposed model for 20k iterations on a single NVIDIA 3090 GPU. We use AdamW optimizer~\cite{loshchilov_2017_decoupled} and the learning rate is $10^{-5}$ during the training. Similar to the baseline, viewpoint images and object bounding boxes, which have been provided by REVERIE, are encoded by ViT-B/16~\cite{dosovitskiy_2020_image} pre-trained on ImageNet~\cite{russakovsky_2015_imagenet}. For the object detection task, we use the Faster R-CNN model~\cite{shaoqing_2015_faster} pre-trained on VG~\cite{krishna_2017_visual}. ConceptNet~\cite{liu_2004_conceptnet} is utilized as the external commonsense knowledge base. To encode object and knowledge labels we employ the pre-trained CLIP model~\cite{radford_2021_Learning}.

\subsection{Dataset and Evaluation Metrics}
REVERIE is a goal-oriented task with concise and high-level instructions that integrates R2R navigation with referring expression grounding. The agent navigates through the environment to identify the referred object that is not visible in the first view. In this dataset, the average length of instructions is 18 words. Also, there are more than 4,000 target objects falling into 489 categories. We utilize the widely-used and standard metrics for performance evaluation. trajectory length (TL), oracle success rate (OSR), success rate (SR), and success rate penalized by path length (SPL) indicate the navigation performance. Moreover, remote grounding success rate (RGS) and remote grounding success rate weighted by path length (RGSPL) relate to object grounding task.

\subsection{Comparison with State-of-the-Arts}
Table~\ref{tab:reverie_results} compares the single-run performance of ACK with state-of-the-art methods on the REVERIE benchmark. Our method achieves state-of-the-art performance and improves all the metrics on test unseen split which demonstrates the effectiveness and generalization ability of our proposed method in unseen environments. According to Table~\ref{tab:reverie_results}, our model remarkably outperforms DUET~\cite{Chen_2022_duet} and consistently enhances all metrics on both validation unseen and test unseen splits. In particular, compared to the baseline model on test unseen split, for navigation metrics, OSR, SR, and SPL improved by $2.10$\%, $1.46$\%, and $1.83$\%, respectively, and for grounding metrics, RGS and RGSPL are improved by $0.89$\% and $1.09$\%, respectively. Note that ACK is just incorporated into the fine-tuning stage of the baseline. Hence, to have a fair comparison, we mentioned the results of KERM~\cite{li_2023_kerm} while its pipeline is only composed of the fine-tuning phase. We also evaluate the performance of ACK on the R2R benchmark, however, no significant improvement is achieved. R2R contains fine-grained and detailed instructions which means the agent can strictly follow instructions, and exploring the environment is not essential. Thus, incorporating commonsense knowledge as complementary data is not very helpful for this task.

\begin{table}[t!]
  \Large
  \begin{center}
  \resizebox{\columnwidth}{!}{
  \begin{tabular}{ccc|ccccc}
    \hline \hline
    
    \multicolumn{1}{c}{KGS} & \multicolumn{1}{c}{CH} & \multicolumn{1}{c|}{CD} & OSR$\uparrow$ & SR$\uparrow$ & SPL$\uparrow$ & RGS$\uparrow$ & RGSPL$\uparrow$ \Tstrut \\
    
    \hline \hline

    $\times$ & $\times$ & $\times$ & 51.01 & 45.92 & 33.77 & 31.30 & 23.31  \\

    $\checkmark$ & $\times$ & $\times$ & 51.36 & 46.18 & 33.81 & 32.04 & 23.33  \\
        
    $\checkmark$ & $\checkmark$ & $\times$ & 52.23 & \textbf{48.08} & 34.02 & 32.25 & 23.39 \\
    
    $\checkmark$ & $\checkmark$ & $\checkmark$ & \textbf{52.77} & 47.49 & \textbf{34.44} & \textbf{32.66} & \textbf{23.92}  \\
    
    \hline \hline
  \end{tabular}}
\end{center}
\caption{Ablation of knowledge graph-aware self-attention, concept history, and commonsense-based decision-making pipeline on REVERIE validation unseen split. The performance is continuously enhanced as the suggested modules are gradually incorporated.}
\label{tab:ablation_kgs_ch_kr}
\end{table}

\begin{table}[t!]
  \begin{center}
  \resizebox{\columnwidth}{!}{
  \begin{tabular}{c|ccccc}
    \hline \hline
    
    \multicolumn{1}{c|}{top-$k$} & OSR$\uparrow$ & SR$\uparrow$ & SPL$\uparrow$ & RGS$\uparrow$ & RGSPL$\uparrow$ \Tstrut \\
    
    \hline \hline
    
    0 & 51.49 & 46.72 & 33.91 & 32.42 & 23.56 \\

    10 & \textbf{52.77} & \textbf{47.49} & \textbf{34.44} & \textbf{32.66} & \textbf{23.92} \\

    20 & 51.66 & 47.25 & 33.98 & 32.49 & 23.67 \\
    
    \hline \hline
  \end{tabular}}
\end{center}
\caption{Ablation of utilizing top-$k$ external knowledge in the training stage on REVERIE validation unseen split.}
\label{tab:ablation_knowledge}
\end{table}

\begin{figure*}[!t]
\begin{center}
\includegraphics[width=0.99\linewidth]{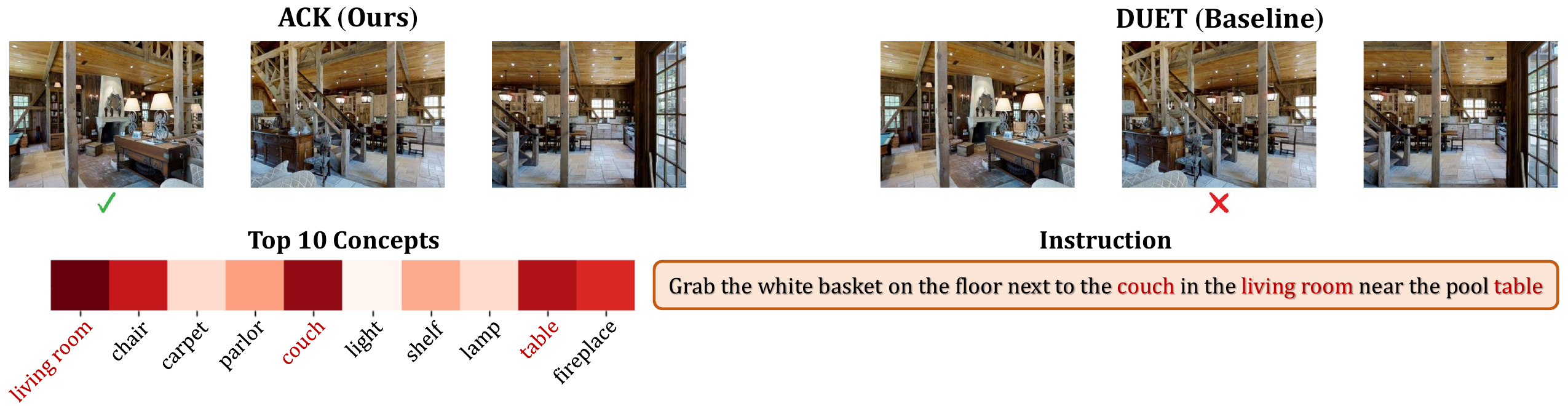}
\end{center}
\caption{Visualization example of navigation performance for comparing ACK and the baseline method. We can see that our method predicts the correct action while DUET selects the wrong candidate direction. The concepts, including detected objects and retrieved commonsense knowledge, with the highest weights are used as landmarks in the instruction. Therefore, taking advantage of these concepts leads to visual representation enhancement and more accurate alignment between visual and textual information. Best viewed in color}
\label{fig:qualitative_analysis}
\end{figure*}

\begin{figure}[!t]
\begin{center}
\includegraphics[width=0.75\linewidth]{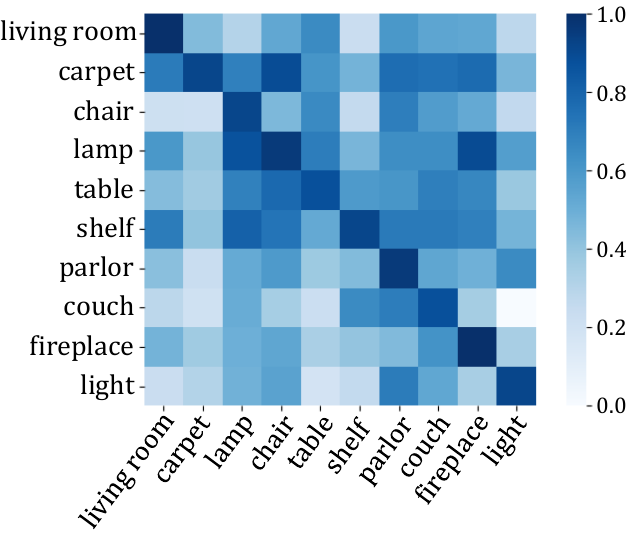}
\end{center}
\caption{Learned concept-to-concept correlation matrix}
\label{fig:c2c_correlation}
\end{figure}

\subsection{Ablation Study}
The contribution of each element is assessed through comprehensive experiments which are shown in Table~\ref{tab:ablation_kgs_ch_kr} and Table~\ref{tab:ablation_knowledge}. The ACK is merely ablated on the validation unseen split of REVERIE.

\noindent\textbf{Knowledge Graph-aware Self-attention.}
As mentioned in Eq.~\ref{eq:kgs} we ablate the transformer with and without knowledge graph structure in Table~\ref{tab:ablation_kgs_ch_kr}. According to this table, if the model is aware of the relationship between concepts, navigation performance is improved.

\noindent\textbf{Concept History.}
Table~\ref{tab:ablation_kgs_ch_kr} ablates the impact of incorporating temporal information of concepts into the proposed model. In REVERIE, it is essential to explore the environment more efficiently due to the lack of step-by-step instructions. Hence, As results in Table~\ref{tab:ablation_kgs_ch_kr} suggest, holding a memory of visited objects in the path alongside the extracted relevant commonsense knowledge helps the agent act more properly in unseen environments.

\noindent\textbf{Commonsense-base Decision-making.}
To show the effectiveness of incorporating commonsense knowledge in the decision-making process, we evaluate the ACK with and without utilizing the new pipeline for action prediction. Table~\ref{tab:ablation_kgs_ch_kr} demonstrates that incorporating the commonsense-based decision-making pipeline into the dynamic fusion module of the baseline makes the agent learn from the knowledge graph and improves local action reasoning.

\noindent\textbf{External Knowledge Capacity.}
To show the impact of commonsense knowledge capacity on agent performance, we evaluate our method by different numbers of knowledge as illustrated in Table~\ref{tab:ablation_knowledge}. Note that, $k=0$ means that we only use the fully-connected scene graph constructed by visible objects in viewpoint images. Increasing the number of knowledge, $k=10$, results in better performance, however, by continuously increasing the number of extracted knowledge, $k=20$, the performance of the agent is deteriorated. It shows extra information may cause noise in the output.

\subsection{Qualitative Analysis}
To visualize our proposed method, we use an example in the validation unseen split of REVERIE. According to the instruction, the target object is the white basket. Figure~\ref{fig:qualitative_analysis} illustrates an example where ACK selects the correct candidate direction by detecting objects and retrieving commonsense knowledge, while the baseline model makes the wrong decision without the supporting facts. In this figure, a heat map is used to visualize the weight distribution over the top 10 concepts of the selected viewpoint, including objects and knowledge, after the knowledge graph-aware cross-modal encoder. Regarding the instruction, \textit{couch} and \textit{table} are used as landmarks. We can see that annotated concepts in red, e.g., \textit{living room}, \textit{couch}, and \textit{table}, can be leveraged as visual landmarks as well as improve the alignment between textual and visual data. Furthermore, the other retrieved concepts such as \textit{chair} and \textit{fireplace}, with higher weights can also be useful for navigation in this example. Also, the learned concept-to-concept correlations for this example are visualized in Figure~\ref{fig:c2c_correlation}.

\section{Conclusion}
In this paper, we propose ACK to enhance visual representation and local action prediction by incorporating commonsense knowledge into the REVERIE task as a spatio-temporal knowledge graph. At first, a knowledge base is constructed and then refined to output purified commonsense information. We further design ACK to leverage the refined commonsense knowledge, which consists of two modules, the knowledge graph-aware cross-modal encoder, and the concept aggregator. The absence of step-by-step instructions in the REVERIE motivated us to hold a concept history during the navigation for more efficient exploration. The experimental results demonstrate the superiority of ACK over the state-of-the-art methods on REVERIE which shows that taking advantage of commonsense knowledge is a promising direction for the REVERIE task. For future work, we aim to exploit other external repositories as well as different formats of commonsense knowledge.

\bibliography{aaai24}

\end{document}